\documentclass{article}

\usepackage[preprint]{nips_2018}

\usepackage[utf8]{inputenc} 
\usepackage[T1]{fontenc}    
\usepackage{hyperref}       
\usepackage{url}            
\usepackage{booktabs}       
\usepackage{amsfonts}       
\usepackage{nicefrac}       
\usepackage{microtype}      

\usepackage{multirow}
\usepackage{graphicx}
\usepackage{epsfig}
\usepackage{algorithm}
\usepackage{algorithmicx}
\usepackage{amsmath,amssymb} 
\usepackage{algpseudocode}

\usepackage{color}

\usepackage{wrapfig}
\usepackage{subfigure}
\usepackage{algcompatible,lipsum}
\usepackage{cases}
\usepackage[skins]{tcolorbox}
\newtcolorbox{mybox}[2][]
  {colback = black, colframe = blue, fonttitle = \bfseries, coltitle = black,
    colbacktitle = white, enhanced,
    attach boxed title to top center={yshift=-3mm},
    title=#2,#1}

\title{Predictive Local Smoothness for Stochastic \\  Gradient Methods}

\author{
  Jun Li$^1$, Hongfu Liu$^1$, Bineng Zhong$^2$, Yue Wu$^1$ and Yun Fu$^1$ \\
  $^1$Department of Electrical and Computer Engineering, Northeastern University, Boston, MA, USA.\\
  $^2$Department of Computer Science and Technology, Huaqiao University, Fujian, China \\
 \texttt{\{junl.mldl,bnzhong\}@gmail.com, liu.hongf@husky.neu.edu} \\
  \texttt{\{yuewu,yunfu\}@ece.neu.edu} \\
}

\begin{document}

\maketitle

\begin{abstract}
Stochastic gradient methods are dominant in nonconvex optimization especially for deep models but have low asymptotical convergence due to the fixed smoothness. To address this problem, we propose a simple yet effective method for improving stochastic gradient methods named predictive local smoothness (PLS). First, we create a convergence condition to build a learning rate which varies adaptively with local smoothness. Second, the local smoothness can be predicted by the latest gradients. Third, we use the adaptive learning rate to update the stochastic gradients for exploring linear convergence rates. By applying the PLS method, we implement new variants of three popular algorithms: PLS-stochastic gradient descent (PLS-SGD), PLS-accelerated SGD (PLS-AccSGD), and PLS-AMSGrad. Moreover, we provide much simpler proofs to ensure their linear convergence. Empirical results show that the variants have better performance gains than the popular algorithms, such as, faster convergence and alleviating explosion and vanish of gradients.
\end{abstract}
\section{Introduction}
In this paper, we consider the following nonconvex optimization:
\begin{align}
\label{eq:problem}
\begin{array}{l@{}l}
\min_{x\in\mathbb{R}^d} f(x):=\frac{1}{n}\sum_{i=1}^nf_i(x),
\end{array}
\end{align}
where $x$ is the model parameter, and neither $f$ nor the individual $f_i$ $(i\in[n])$ are convex, such as, deep models.
Stochastic gradient descent (SGD) is one of the most popular algorithms for minimizing the loss function in Eq. \eqref{eq:problem}. It iteratively updates the parameter by using the product of a learning rate and the negative gradient of the loss, which is computed on a minibatch drawn randomly from training set. Unfortunately, small learning rate makes SGD painfully slow to converge, while high learning rate causes SGD to diverge. Therefore, choosing a proper learning rate becomes a challenge.

Recently, an adaptive method adjusts automatically the learning rate by using some forms of the past gradients to scale coordinates of the gradient. AdaGrad \cite{Duchi2011AdaGrad} is the first popular adaptive algorithm to update the sparse gradients by dividing positive square root of averaging the squared past gradients. However, its implementation leads to rapid decay of the learning rate for dense gradients. To address this issue, there are several variants of AdaGrad (e.g., Adadelta \cite{Zeiler2012Adadelta}, RMSProp \cite{Tieleman2012RmsProp}, Adam \cite{Kingma2015Adam}, Nadam \cite{Dozat2016Nadam}, and AMSGrad \cite{Reddi2018AMSGrad}), which have been widely and successfully applied to train deep models. Especially, they use the exponential moving averages of squared past gradients to manage the rapidly decayed learning rate. In addition, momentum is another method to keep velocity of the gradients for passing through trouble navigating ravines of the loss function (e.g., heavy ball (HB) \cite{Polyak1964}, Nesterov's accelerated gradient descent (NAG) \cite{Nesterov1983}, and accelerated SGD (AccSGD) \cite{Jain2017accSGD,Kidambi2018accSGD}).

\begin{figure}[t]
\vskip -0.0in
\begin{center}
\centerline{\includegraphics[width=1\columnwidth]{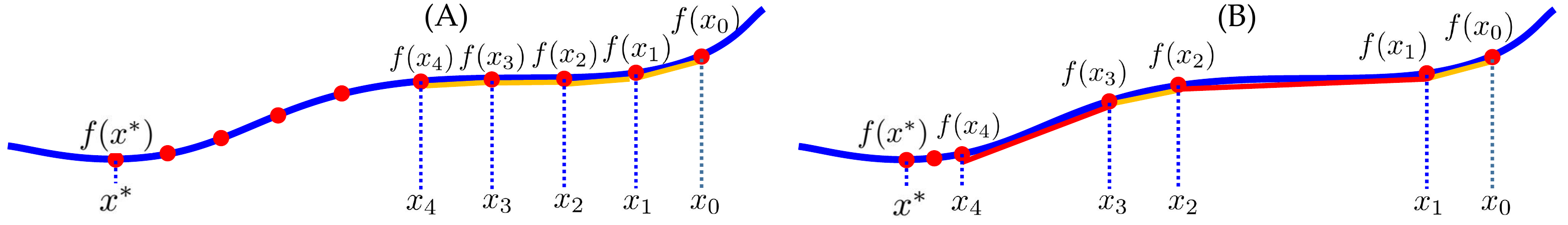}}
\vskip -0.1in
\caption{(A) Maximum smoothness results in low learning rate to slowly move from an initial point $f(x_0)$ to local minimum $f(x^\ast)$, where the blue line denotes the function and the orange line is the tangent. (B) Predictive local smoothness leads to adaptive learning rate for fast moving from $f(x_0)$ to $f(x^\ast)$. The red and orange lines correspond to large and small learning rates, respectively.}
\label{fig:smoothnesslabel}
\end{center}
\vskip -0.25in
\end{figure}

However, establishing convergence guarantees for above mentioned methods are based on a maximum of Lipschitz constant $L$ which masters smoothness of the loss function in whole parameter space, called \emph{$L$-smoothness} \cite{Bottou2016omml}. Since the learning rate (step size) is inversely proportional to the $L$-smoothness \cite{Bottou2016omml}, the maximum $L$ results in a low learning rate to slowly move the loss function from a point $f(x_0)$ to local minimum loss $f(x^\ast)$ with an equilibrium parameter $x^\ast$ in Figure \ref{fig:smoothnesslabel} (A) in the popular algorithms (e.g., SGD, AMSGrad \cite{Reddi2018AMSGrad} and AccSGD \cite{Kidambi2018accSGD}). In fact, the learning rate adaptively varies on the updating parameter in Figure.\ref{fig:smoothnesslabel} (B) because it truly depends on a local smoothness between the current parameter $x_t$ and the equilibrium parameter $x^\ast$. Naturally, the local smoothness leads to high learning rate for fast decreasing the loss values as it is lower than the maximum $L$. Thus, this arouses us to choose the learning rate by using the local smoothness between $x_t$ and $x^\ast$.

This paper, therefore, provides a local smoothness strategy to study the adaptive learning rate. In this strategy there are two important problems that how to easily build a direct functional relationship between the learning rate and the local smoothness, and how to calculate the local smoothness. To issue these problems, the stochastic gradient algorithms are transformed into a linear dynamical system by using the local smoothness to linearize the gradient function. The functional relationship is obtained by constructing the convergence condition for the linear dynamical system. Although the local smoothness between $x_t$ and $x^\ast$ is not easily calculated due to the unknown equilibrium $x^\ast$, it is simply predicted by using the latest gradients. Overall, our main contributions are summarized as follows:
\begin{itemize}
\item We propose a predictive local smoothness (PLS) method to adjust automatically learning rate for stochastic gradient algorithms. Our ALS method will lead these algorithms to drive a loss function to fast converge to a local minimum.
\item We apply PLS into SGD, the classical adaptive method based on exponential moving averages, AMSGrad \cite{Reddi2018AMSGrad}, and the typical momentum method, AccSGD \cite{Kidambi2018accSGD}. Correspondingly, we establish three PLS-SGD, PLS-AMSGrad and PLS-AccSGD algorithms, and provide corresponding theoretical conditions to ensure their linear convergence.
\item We also provide an important empirical result that PLS can alleviate the exploding and vanishing gradients in the classical algorithms (e.g., SGD, AMSGrad \cite{Reddi2018AMSGrad} and AccSGD \cite{Kidambi2018accSGD}) for training deep model with least squares regression loss and rectified linear units (ReLU).
\end{itemize}
\section{Preliminaries}
In this section, we introduce some notations, local smoothness assumption, and three popular stochastic gradient algorithms: SGD, AMSGrad \cite{Reddi2018AMSGrad}, and AccSGD \cite{Kidambi2018accSGD}.

\textbf{Notation.} $\nabla f(x)$ denotes exact gradient of $f$ at $x$, while $\nabla f_{i_t}(x)$ denotes a stochastic gradient of $f$, where $i_t$ is sampled uniformly at random from $[1,\cdots ,n]$ and $n$ is the number of samples. Since $i_t$ is sampled in an independent and identically distributed (IID) manner from $\{1,\cdots,n\}$, the expectation of smoothness $L_{i_t}(x_t)$ is denoted as $\mathbb{L}(x_t)=\mathbb{E}\left[L_{i_t}(x_t)\right]=\frac{1}{n}\sum_{i=1}^nL_{i}(x_t)$. A $\ell_1$ or $\ell_2$-norm of a vector $x$ is denoted as $\|x\|$, and its square is $\|x\|^2$. A positive-definite matrix $A$ is denoted as $A\succ 0$. The Kronecker product of matrices $A$ and $B$ is denoted as $A\otimes B$. We denote a $d\times d$ identity matrix by $I_d$. A neighborhood of a point $x_1\in\mathbb{R}^d$ with radius $r_{x_2}$ is denoted as $N_{r_{x_2}}(x_1)=\{y\in\mathbb{R}^d| \|x_1-y\|<r_{x_2}=\|x_1-x_2\|\}$.

\textbf{Assumption 1.} \emph{We say $f$ is local smoothness on a set $C \subset\mathbb{R}^d$ if there is a constant $L$ such that
\begin{align}
\label{eq:lipschitz0}
\|\nabla f(x)-\nabla f(y)\|\leq L\|x-y\|, \ \ \ \text{for}\ \ \forall\ x,y \in C.
\end{align}}Assumption 1 is an essential foundation for convergence guarantees of most stochastic gradient methods as the gradient of $f$ is controlled by $L$ with respect to the parameter vector. 

\textbf{SGD.} Stochastic Gradient Descent (SGD) simply computes the gradient of the parameters by uniformly randomly choosing a single or a few training examples. Its update is given by
\begin{align}
\label{eq:sgd}
x_{t+1}=x_{t}-\eta_t\nabla f_{i_t}(x_{t}),
\end{align}
where $\eta_t$ is the learning rate. $\eta_t$ is usually set to a decay form $\eta_0/\sqrt{t}$ in practice. This setting leads to slower convergence. Moreover, the gradient of the loss of a deep model with rectified linear units (ReLU) \cite{Nair2010relu} often explodes when large initialization $\eta_0$.

\textbf{AMSGrad.} AMSGrad \cite{Reddi2018AMSGrad} is an exponential moving average variant of the popular Adam algorithm \cite{Kingma2015Adam} in the scale gradient method. AMSGrad uses the factors $\beta_{1t}=\beta_{1}/t$ and $\beta_{2}$ to exponentially move the momentums of the gradient and the squared gradient, respectively. $\beta_{1t}=\beta_{1}=0.9$ and $\beta_{2}=0.999$ are typically recommended in practice. The key update is described as follows:
\begin{align}
\label{eq:AMSGrad}
\left\{\begin{array}{ccc}
m_{t+1}=\beta_{1t}m_{t}+(1-\beta_{1t})\nabla f_{i_t}(x_{t}), \ \ \ \ \ \ \ \ \ \ \ \ \ \ \ \ \ \ \ \ \  \\
v_{t+1}=\beta_{2}v_{t}+(1-\beta_{2})\left(\nabla f_{i_t}(x_{t})\right)^2, \ \ \ \ \ \ \ \ \ \ \ \ \ \ \ \  \  \\
\widehat{v}_{t+1}=\max\{v_{t+1},\widehat{v}_{t}\}, \ \ \ x_{t+1}=x_{t}-\eta_t\frac{m_{t+1}}{\sqrt{\widehat{v}_{t+1}}}.
\end{array}
\right.
\end{align}
\textbf{AccSGD.} Accelerated SGD (AccSGD) proposed in \cite{Jain2017accSGD} is much better than SGD, HB \cite{Polyak1964} and NAG \cite{Nesterov1983} in the momentum method. An intuitive version of AccSGD is presented in \cite{Kidambi2018accSGD}. Particularly, AccSGD takes three parameters: learning rate $\eta_t$, long learning rate parameter $\kappa\geq 1$, statistical advantage parameter $\xi\leq\sqrt{\kappa}$, and $\alpha=1-0.7^2\xi/\kappa$. This update can alternatively be stated by:
\begin{align}
\label{eq:AccSGD}
\left\{\begin{array}{ccc}
m_{t+1}=\alpha m_{t}+(1-\alpha)\left(x_{t}-\frac{\kappa\eta_t}{0.7}\nabla f_{i_t}(x_{t})\right),\ \ \ \ \ \ \ \ \ \ \ \ \ \ \ \ \ \ \  \\
x_{t+1}=\frac{0.7}{0.7+(1-\alpha)}\left(x_{t}-\eta_t\nabla f_{i_t}(x_{t})\right)+\frac{1-\alpha}{0.7+(1-\alpha)}m_{t+1}.
\end{array}
\right.
\end{align}
\section{Predictive Local Smoothness}
The popular adaptive learning rate methods are based on using gradient updates scaled by square roots of exponential moving averages of squared past gradients \cite{Reddi2018AMSGrad}. These methods indirectly adjust the learning rate as they can be essentially viewed as gradient normalization. In this section, we study the local smoothness to directly and adaptively adjust the learning rate, propose a predictive local smoothness (PLS) method, and apply this method into SGD, AMSGrad \cite{Reddi2018AMSGrad}, and AccSGD \cite{Kidambi2018accSGD}. Before showing our PLS method, we first give a local smoothness sequence definition.

\textbf{Definition 1.} \emph{Let $x^\ast$ be an equilibrium of the local minimum $f(x^\ast)$ and $\{x_t\}_{t\geq 0}$ be a updating parameter procedure, where $x_0$ is an initial point. A corresponding neighborhood sequence of $x^\ast$ is denoted by $\{N_{r_{x_t}}(x^\ast)\}_{t\geq 0}$, where $r_{x_t}=\|x^\ast-x_t\|$. A \textbf{local smoothness sequence} of $x^\ast$ is defined as $\{\mathcal{L}(x_t)\}_{t\geq 0}$ which satisfies that,
\begin{align}
\label{eq:lipschitz}
\|\nabla f(x^\ast)-\nabla f(y)\|\leq \mathcal{L}(x_t)\|x^\ast-y\|,\ \  \text{for} \ \ \forall\ y \in N_{r_{x_t}}(x^\ast).
\end{align}
A forward neighborhood sequence on $\{x_t\}_{t\geq 1}$ is denoted by $\{N_{r_{x_{t+1}}}(x_t)\}_{t\geq 1}$, where $r_{x_{t+1}}=\|x_{t+1}-x_{t}\|$. An \textbf{ideal local smoothness sequence} is defined as $\{\overline{L}(x_t)\}_{t\geq 0}$ which satisfies that
\begin{align}
\label{eq:ideallipschitz}
\|\nabla f(x_t)-\nabla f(y)\|\leq \overline{L}(x_t)\|x_t-y\|,\ \  \text{for} \ \ \forall\ y \in N_{r_{x_{t+1}}}(x_t).
\end{align}
A backward neighborhood sequence on $\{x_t\}_{t\geq 1}$ is denoted by $\{N_{r_{x_{t-1}}}(x_t)\}_{t\geq 1}$, where $r_{x_{t-1}}=\|x_t-x_{t-1}\|$. A \textbf{predictive local smoothness sequence} is defined as $\{L(x_t)\}_{t\geq 1}$ which satisfies that
\begin{align}
\label{eq:predictlipschitz}
\|\nabla f(x_t)-\nabla f(y)\|\leq L(x_t)\|x_t-y\|,\ \  \text{for} \ \ \forall\ y \in N_{r_{x_{t-1}}}(x_t).
\end{align}}This definition reveals three local smoothness sequences. $\mathcal{L}(x_t)$ masters the smoothness between the equilibrium $x^\ast$ and $x_t$ to strictly ensure the convergence of the updating parameter procedure. $\overline{L}(x_t)$ is an ideal local smoothness to fast move $x_t$ to $x_{t+1}$. Both $\mathcal{L}(x_t)$ and $\overline{L}(x_t)$ cannot be computed due to the unknown $x^\ast$ and $x_{t+1}$, while $L(x_t)$ is easily calculated by using $x_{t}$ and $x_{t-1}$.
\subsection{PLS method}
PLS is a new adaptive learning rate method based on the local smoothness. The local smoothness varies with the updating parameters $\{x_t\}_{t\geq 0}$ in stochastic gradient algorithms. Based on the above definition, the local smoothness sequence is $\{\mathcal{L}(x_t)\}_{t\geq 0}$, which will be used to adjust the learning rate, in the neighborhood sequence $\{N_{r_{x_t}}(x^\ast)\}_{t\geq 0}$. However, it is difficult to compute $\{\mathcal{L}(x_t)\}_{t\geq 0}$ because of the unknown $x^\ast$. Although $\mathcal{L}(x_t)$ cannot be calculated, it can be predicted by $L(x_t)$. Fortunately, the sequence $\{L(x_t)\}_{t\geq 1}$ is easily predicted by using the current gradient and the latest gradient in the neighborhood sequence $\{N_{r_{x_{t-1}}}(x_t)\}_{t\geq 1}$. In this paper, our key idea is to use the predictive local smoothness sequence $\{L(x_t)\}_{t\geq 1}$ instead of the unknown $\{\mathcal{L}(x_t)\}_{t\geq 0}$ to adjust automatically the learning rate $\eta_t$. PLS is described as the following three steps.
\begin{figure}[t]
\vskip -0.0in
\begin{center}
\centerline{\includegraphics[width=1\columnwidth]{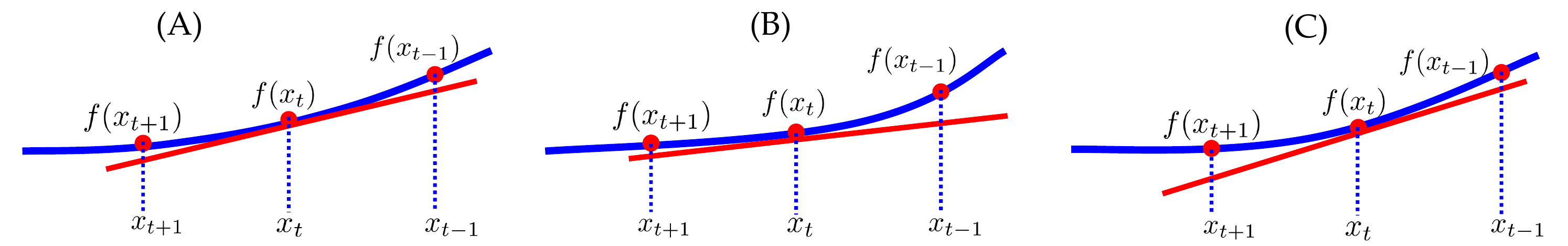}}
\vskip -0.1in
\caption{Three cases between the ideal $\overline{L}(x_t)$ and the predictive $L(x_t)$, which is predicted by Eq. \eqref{eq:predictlipschitz}. (A) $L(x_t)=\overline{L}(x_t)$. (B) $L(x_t)>\overline{L}(x_t)$. (C) $L(x_t)<\overline{L}(x_t)$. }
\label{fig:predictivesmoothness}
\end{center}
\vskip -0.2in
\end{figure}

\textbf{Building adaptive learning rate with local smoothness.} We first create a functional relationship between $\eta_t$ and $L(x_t)$ by using the convergence conditions of stochastic gradient algorithms. Following the local smoothness in Eq. \eqref{eq:lipschitz}, since $\nabla f(x^\ast)=0$, $\nabla f(x_t)$ is linearized by $\nabla f(x_t)= $ $\left(\mathcal{L}(x_t)\otimes I_d\right)(x_t-x^\ast)$. Using $L(x_t)$ instead of $\mathcal{L}(x_t)$, we consider the following linearization
\begin{align}
\label{eq:linearizing}
\nabla f(x_t)= \left(L(x_t)\otimes I_d\right)(x_t-x^\ast),
\end{align}
where $L(x_t)$ is computed by Eq. \eqref{eq:predictlipschitz} in the neighborhood $N_{r_{x_{t-1}}}(x_t)$, and $L(x_t)=\|\nabla^2 f(x_t)\|$ if $f$ is twice continuously differentiable. Stochastic gradient algorithms can use the linearization Eq. \eqref{eq:linearizing} to transform it into a simple time-varying linear system. The convergence of the algorithms is achieved by studying the stability of this linear system \cite{Lessard2016iqc}. Therefore, the stability condition is naturally used to construct the functional relationship between $\eta_t$ and $L(x_t)$, $\eta_t=\eta(L(x_t))$. This shows that the learning rate is adaptively tuned by $L(x_t)$.

\textbf{Predicting the local smoothness.} We secondly predict the local Lipschitz constant $L(x_t)$ by using the current gradient $\nabla f(x_t)$ and the latest gradient $\nabla f(x_{t-1})$. By using the local smoothness in Eq. \eqref{eq:lipschitz}, $L(x_t)$ on $N_{r_{x_{t-1}}}(x_t)$ is predicted by
\begin{align}
\label{eq:predictlipschitz}
L(x_t)=\frac{\|\nabla f(x_t)-\nabla f(x_{t-1})\|}{\|x_t-x_{t-1}\|+\epsilon_1},
\end{align}
where $\epsilon_1$ is a parameter to prevent $\|x_t-x_{t-1}\|$ going to zero. This predictive Lipschitz constant $L(x_t)$ is utilized to adjust automatically the learning rate $\eta_t$ for computing the parameter $x_{t+1}$. In the next subsections, we prove that $\eta_t$ is inversely proportional to $L(x_t)$, $\eta_t\propto (1/(L(x_t)+\epsilon_2))$, where $\epsilon_2$ is another parameter to avoid the learning rate to be over large in the later updating process. Next, we discuss the relationship between the predictive $L(x_t)$ and the ideal $\overline{L}(x_t)$ for the learning rate $\eta_t$.

The key question is whether the predictive $L(x_t)$ is fit for the ideal $\overline{L}(x_t)$. There are three cases in Figure \ref{fig:predictivesmoothness}. Case 1) in Figure \ref{fig:predictivesmoothness}(A): $L(x_t)$ is similar or equal to $\overline{L}(x_t)$, which shows a prefect predictive $L(x_t)$ for calculating the parameter $x_{t+1}$. Case 2) in Figure \ref{fig:predictivesmoothness}(B): $L(x_t)$ is greater than $\overline{L}(x_t)$. It reveals the predictive $L(x_t)$ can be used to compute the parameter $x_{t+1}$ although it reduces the learning rate as $\eta_t\propto (1/L(x_t))$. Case 3) in Figure \ref{fig:predictivesmoothness}(C): $L(x_t)$ is less than $\overline{L}(x_t)$. This is an ill prediction as the less $L(x_t)$ leads to a bigger learning rate and a larger loss. Fortunately, it spends some time to decrease the larger loss since it will be decreased by stochastic gradient descent algorithms in practice.

\textbf{Applying the adaptive learning rate into any stochastic gradient algorithms.}
We thirdly use the adaptive learning rate $\eta_t=\eta(L(x_t))$ to update the parameter $x_{t+1}$ in the stochastic gradient algorithms. Overall, Figure \ref{fig:procedurepls} summarizes the proposed predictive local smoothness method. This method can be applied into the adaptive method based on exponential moving averages, for example, AdaGrad \cite{Duchi2011AdaGrad}, Adadelta \cite{Zeiler2012Adadelta}, RMSProp \cite{Tieleman2012RmsProp}, Adam \cite{Kingma2015Adam}, Nadam \cite{Dozat2016Nadam} and AMSGrad \cite{Reddi2018AMSGrad}, and the momentum methods, such as, HB \cite{Polyak1964}, NAG \cite{Nesterov1983} and AccSGD \cite{Jain2017accSGD,Kidambi2018accSGD}. Next, we will apply PLS into SGD, AMSGrad \cite{Reddi2018AMSGrad} and AccSGD \cite{Kidambi2018accSGD} to show its effectiveness.

\begin{figure}[t]
\begin{mybox}[colback=white]{Procedure PLS}
\begin{itemize}
\item Build adaptive learning rate with local smoothness $\eta=\eta(L(x_t))$ by using Eq. \eqref{eq:predictlipschitz}.
\item Predict local smoothness $L(x_t)=\frac{\|\nabla f(x_t)-\nabla f(x_{t-1})\|}{\|x_t-x_{t-1}\|+\epsilon_1}$.
\item Apply $\eta(L(x_t))\propto\frac{1}{L(x_t)+\epsilon_2}$ to update $x_{t+1}$ by any stochastic gradient algorithms.
\end{itemize}
\end{mybox}
\vskip -0.1in
\caption{Predictive Local Smoothness.}
\label{fig:procedurepls}
\vskip -0.1in
\end{figure}

\textbf{Remark 1.} \emph{Compared the related local smoothness methods \cite{Kpotufe2013localsmoothKernelR,Vainsencher2015localsvrg}, they require that the loss function is twice continuously differentiable, and are applied into the kernel regression and stochastic variance reduced gradient (SVRG) \cite{Johnson2013svrg,Reddi2016svrno}, while we need the loss function with continuously differentiable and mainly study the stochastic gradient algorithms (e.g., SGD, AMSGrad \cite{Reddi2018AMSGrad} and AccSGD \cite{Kidambi2018accSGD}). In addition, the unknown global smoothness is estimated along with the optimization \cite{Malherbe2017GOLF}, while we provide an effective method to predict unknown local smoothness using Eq. \eqref{eq:predictlipschitz}.}

\subsection{PLS-SGD}
\begin{wrapfigure}[13]{r}{2.5in}
\vskip -0.3in
\begin{minipage}{.46\textwidth}
\begin{algorithm}[H]
\caption{PLS-SGD.}
\begin{algorithmic}[1]
\State \textbf{input:} $\eta_0$, $\epsilon_1$, $\epsilon_2$.
\State \textbf{initialize:} $x_0$.
\State \textbf{for} $t=1,\cdots, T-1$ \textbf{do}
\State \hspace{0.4cm} Randomly pick $i_t$ from $\{1,\cdots,n\}$;
\State \hspace{0.4cm} $g_{t}=\nabla f_{i_t}(x_{t})$;
\State \hspace{0.4cm} $L_{i_t}(x_t)=\frac{\|g_{t}-g_{t-1}\|}{\|x_t-x_{t-1}\|+\epsilon_1}$;
\State \hspace{0.4cm} $\eta_t=\frac{\eta_0}{(L_{i_t}(x_t)+\epsilon_2)}$;
\State \hspace{0.4cm} $x_{t+1}=x_{t}-\eta_t\nabla f_{i_t}(x_{t})$;
\State \textbf{end for}
\end{algorithmic}
\label{alg:PLS-SGD}
\end{algorithm}
\end{minipage}
\end{wrapfigure}
In this subsection, we introduce a PLS-SGD algorithm. By using the linearization Eq. \eqref{eq:linearizing} in the PLS method and computing the expectation, the updating rule of \textbf{SGD} is converted into the linear system:
\begin{small}
\begin{align}
\label{eq:linearSGD}
x_{t+1}-x^\ast=\left((1-\eta_t\mathbb{L}(x_t))\otimes I_d\right)(x_{t}-x^\ast),
\end{align}
\end{small}
where $\mathbb{L}(x_t)=\mathbb{E}\left[L_{i_t}(x_t)\right]=\frac{1}{n}\sum_{i=1}^nL_{i}(x_t)$ as $i_t$ is sampled in an IID manner from $\{1,\cdots,n\}$.
Then, the convergence condition of SGD is obtained by employing the stability condition of the linear system in Eq. \eqref{eq:linearSGD}, which shows that $x_{t}$ converges to $x^\star$ at a given linear rate $\rho$. Now we present a linear convergence condition for the SGD as follows.

\textbf{Theorem 1\footnote{All the proofs for the Theorems and the linear systems are provided in the Appendix.}.} \emph{Consider the linear system in Eq. \eqref{eq:linearSGD}. Assume that $i_t$ is sampled in an IID manner from a uniform distribution, the assumption 1 holds and there exists an equilibrium $x^\ast\in\mathbb{R}^d$ such that $\nabla f(x^\ast)=0$. For a fixed linear convergence rate $0<\rho<1$, if $(1-\rho)\frac{1}{\mathbb{L}(x_t)}\leq \eta_t\leq \frac{1}{\mathbb{L}(x_t)}$ holds, then the linear system is exponentially stable, that is, $\|x_t-x^\ast\|_2\leq \rho^t \|x_0-x^\ast\|_2$.}

Theorem 1 provides a condition $0<1-\eta_t\mathbb{L}(x_t)\leq \rho$ for the linear convergence of SGD, which benefits from our PLS method. The condition revePLS that the functional relationship between $\eta_t$ and $\mathbb{L}(x_t)$ is $\eta_t=\eta_0/\mathbb{L}(x_t)$, where $\eta_0$ is an initialized learning rate and $1-\rho\leq\eta_0\leq1$. $\mathbb{L}(x_t)$ is sampled uniformly at random from $[1,\cdots ,n]$, that is, $L_{i_t}(x_t)$, which is predicted by using Eq. \eqref{eq:predictlipschitz}. Similar to $\epsilon_1$, $\epsilon_2$ is another parameter to stop $L_{i_t}(x_t)$ going to zero, and avoid the learning rate to be over large in the latter updating process. In our PLS-SGD algorithm, therefore, the learning rate $\eta_t$ is set to $\eta_0/(L_{i_t}(x_t)+\epsilon_2)$. The adaptive learning rate results in that PLS-SGD has a faster (linear) convergence rate than the traditional SGD. PLS-SGD is summarized in \textbf{Algorithm \ref{alg:PLS-SGD}}.

\subsection{PLS-AMSGrad}
Similar to PLS-SGD, we integrate the proposed PLS method into the classical adaptive method based on exponential moving averages, AMSGrad \cite{Reddi2018AMSGrad}, and propose a PLS-AMSGrad algorithm. The Lipschitz linearization Eq. \eqref{eq:linearizing} is used to linearize the updating rules \eqref{eq:AMSGrad} of AMSGrad as:
\begin{align}
\label{eq:linearAMSGrad}
\left(\hspace{-0.2cm}
\begin{array}{cc}
m_{t+1} \\
x_{t+1}-x^\ast \\
\end{array}\hspace{-0.2cm}
\right)=\left(A_t\otimes I_d\right)\left(\hspace{-0.2cm}
\begin{array}{cc}
m_{t} \\
x_{t}-x^\ast \\
\end{array}\hspace{-0.2cm}
\right), \ \ \ \text{where}\ \ \  A_t=\left(\hspace{-0.2cm}
\begin{array}{cc}
\beta_{1t} & (1-\beta_{1t})\mathbb{L}(x_t)\\
-\frac{\eta_t\beta_{1t}}{\sqrt{\widehat{v}_{t+1}}}& 1-\frac{(1-\beta_{1t})\eta_t\mathbb{L}(x_t)}{\sqrt{\widehat{v}_{t+1}}}\\
\end{array}\hspace{-0.2cm}
\right),
\end{align}
$\widehat{v}_{t+1}=\max\{v_{t+1},\widehat{v}_{t}\}$, $v_{t+1}=\left(\beta_{2}\otimes I_d\right)v_{t}+\left((1-\beta_{2})\mathbb{L}^2(x_t)\otimes I_d\right)(x_{t}-x^\ast)^2$ and $\mathbb{L}(x_t)$ is defined in Eq. \eqref{eq:linearSGD}. In fact, $m_t$ is the momentum method to manage the velocity of the gradient. Using this linearization, we provide a much simpler convergence analysis of AMSGrad by studying the linear system in Eq. \eqref{eq:linearAMSGrad}. The linear convergence condition of AMSGrad is described as follows.

\textbf{Theorem 2.} \emph{Consider the linear system in Eq. \eqref{eq:linearAMSGrad}. Assume that $i_t$ is sampled in an IID manner from a uniform distribution, the assumption 1 holds and there exists an equilibrium $x^\ast\in\mathbb{R}^d$ such that $\nabla f(x^\ast)=0$. For a fixed linear convergence rate $\rho=\max_t\{\sqrt{\beta_{1t}}\}$, if there exists a $2\times2$ positive definite matrix $P\succ 0$ such that
\begin{align}
\label{eq:condition1AMSGrad}
A_t^TPA_t-\rho^2 P\prec0,
\end{align} or the following condition holds
\begin{align}
\label{eq:condition2AMSGrad}
\frac{\left(1-\sqrt{\beta_{1t}}\right)\sqrt{\widehat{v}_{t+1}}}{1+\sqrt{\beta_{1t}}}\frac{1}{\mathbb{L}(x_t)}<\eta_t<
\frac{\left(1+\sqrt{\beta_{1t}}\right)\sqrt{\widehat{v}_{t+1}}}{1-\sqrt{\beta_{1t}}}\frac{1}{\mathbb{L}(x_t)},
\end{align}
then the linear system is exponentially stable, that is, $\left\|\hspace{-0.2cm}
\begin{array}{cc}
m_{t+1} \\
x_{t+1}-x^\ast \\
\end{array}\hspace{-0.2cm}\right\|_2\leq \sqrt{\text{cond}(P)} \rho^t \left\|\hspace{-0.2cm}
\begin{array}{cc}
m_{0} \\
x_{0}-x^\ast \\
\end{array}\hspace{-0.2cm}\right\|_2$, where $\text{cond}(P)$ is the condition number of $P$ and $\text{Cond}(A)=\sigma_1(A)/\sigma_p(A)$, where $\sigma_1(A)$ and $\sigma_p(A)$ denote the largest and smallest singular values of the matrix $A$.}

\begin{wrapfigure}[15]{r}{2.5in}
\vskip -0.3in
\begin{minipage}{.46\textwidth}
\begin{algorithm}[H]
\caption{PLS-AMSGrad.}
\begin{algorithmic}[1]
\State \textbf{input:} $\eta_0>0$, $\{\beta_{1t}>0\}_{t=0}^{T-1}$, $\beta_{2}$, $\epsilon_1$, $\epsilon_2$.
\State \textbf{initialize:} $x_0=0$, $u_0=v_0=\widehat{v}_0=0$.
\State \textbf{for} $t=1,\cdots, T-1$ \textbf{do}
\State \hspace{0.4cm} Randomly pick $i_t$ from $\{1,\cdots,n\}$;
\State \hspace{0.4cm} $g_{t}=\nabla f_{i_t}(x_{t})$;
\State \hspace{0.4cm} $L_{i_t}(x_t)=\frac{\|g_{t}-g_{t-1}\|}{\|x_t-x_{t-1}\|+\epsilon_1}$;
\State \hspace{0.4cm} $\eta_t=\frac{\eta_0}{(L_{i_t}(x_t)+\epsilon_2)}$;
\State \hspace{0.4cm} $m_{t+1}=\beta_{1t}m_{t}+(1-\beta_{1t})g_{t}$;
\State \hspace{0.4cm} $v_{t+1}=\beta_{2}v_{t}+(1-\beta_{2})g_{t}^2$;
\State \hspace{0.4cm} $\widehat{v}_{t+1}=\max\{v_{t+1},\widehat{v}_{t}\}$;
\State \hspace{0.4cm} $x_{t+1}=x_{t}-\eta_t\frac{m_{t+1}}{\sqrt{\widehat{v}_{t+1}}}$;
\State \textbf{end for}
\end{algorithmic}
\label{alg:PLS-AMSGrad}
\end{algorithm}
\end{minipage}
\end{wrapfigure}
Compared to the convergence analysis of AMSGrad in \cite{Reddi2018AMSGrad}, Theorem 2 establishes simpler conditions \eqref{eq:condition1AMSGrad} and \eqref{eq:condition2AMSGrad} for its linear convergence. The $2\times2$ linear matrix inequality (LMI) condition \eqref{eq:condition1AMSGrad} is built by using the control theory (e.g., integral quadratic constraint \cite{Lessard2016iqc}) to study the stability of the linear system  \eqref{eq:linearAMSGrad}. It is easily solved by LMI toolbox \cite{Boyd1994lmi}. Although the condition \eqref{eq:condition1AMSGrad} is not very clear to the relationship between $\eta_t$ and $\mathbb{L}(x_t)$, the condition \eqref{eq:condition1AMSGrad} directly reveals its functional relationship, that is, $\eta_t=\overline{\eta}_t\sqrt{\widehat{v}_{t+1}}/\mathbb{L}(x_t)$, where $\frac{1-\sqrt{\beta_{1t}}}{1+\sqrt{\beta_{1t}}}<\overline{\eta}_t<
\frac{1+\sqrt{\beta_{1t}}}{1-\sqrt{\beta_{1t}}}$. Based on Eq. \eqref{eq:AMSGrad}, $\widehat{v}_{t+1}$ tends to zero as it is a linear system, $0<\beta_2<1$, and the gradient goes to zero. For simplification, $\eta_t=\eta_0/\mathbb{L}(x_t)$, where $\eta_0$ is an initialized learning rate and $\frac{1-\sqrt{\beta_{1}}}{1+\sqrt{\beta_{1}}}<\eta_0<
\frac{1+\sqrt{\beta_{1}}}{1-\sqrt{\beta_{1}}}$, since $\beta_{1t}=\beta_{1}$. Similar to PLS-SGD, $\mathbb{L}(x_t)$ is also sampled uniformly at random from $[1,\cdots ,n]$, that is, $L_{i_t}(x_t)$ is computed by Eq. \eqref{eq:predictlipschitz}, and the learning rate $\eta_t$ is set to $\frac{\eta_0}{(L_{i_t}(x_t)+\epsilon_2)}$ or $\frac{\eta_0}{\sqrt{t}(L_{i_t}(x_t)+\epsilon_2)}$ for avoiding the over-large learning rate in the latter updating process. Thus, PLS-AMSGrad is summarized in \textbf{Algorithm \ref{alg:PLS-AMSGrad}}. 

\textbf{Remark 2.} \emph{Based on the Lemma, the $2\times2$ LMI in Eq. \eqref{eq:condition1AMSGrad} is equivalent to the condition in Eq. \eqref{eq:condition2AMSGrad}. The former is obtained by constructing the Lyapunov function in the control theory, while the latter is built by calculating the spectral radius of the weight matrix in the linear system in Eq. \eqref{eq:linearAMSGrad}, which is defined as the magnitude of the largest eigenvalue of the weight matrix. }

\subsection{PLS-AccSGD}

In this subsection, we present a PLS-AccSGD algorithm. Similar to PLS-AMSGrad, our PLS method is integrated into the classical momentum method, AccSGD \cite{Jain2017accSGD,Kidambi2018accSGD}. Using the Lipschitz linearization Eq. \eqref{eq:linearizing}, the updating rules in Eq. \eqref{eq:AccSGD} of AccSGD is simply linearized as:
\begin{align}
\label{eq:linearAccSGD}
\left(\hspace{-0.2cm}
\begin{array}{cc}
m_{t+1}-x^\ast \\
x_{t+1}-x^\ast \\
\end{array}\hspace{-0.2cm}
\right)=\left(B_t\otimes I_d\right)\left(\hspace{-0.2cm}
\begin{array}{cc}
m_{t}-x^\ast\\
x_{t}-x^\ast \\
\end{array}\hspace{-0.2cm}
\right),
\end{align}
where $B_t =\left(\hspace{-0.2cm}
\begin{array}{cc}
\alpha & (1-\alpha)(1-a\eta_t \mathbb{L}(x_t))\\
b\alpha & (1-b)(1-\eta_t\mathbb{L}(x_t))+b(1-\alpha)(1-a\eta_t \mathbb{L}(x_t))\\
\end{array}\hspace{-0.2cm}
\right)$, $\alpha=1-\frac{0.7^2\xi}{\kappa}$, $a=\frac{\kappa}{0.7}$, $b=\frac{1-\alpha}{0.7+(1-\alpha)}$, $\xi$ and $\kappa$ are defined in Eq. \eqref{eq:AccSGD}, and $\mathbb{L}(x_t)$ is defined in Eq. \eqref{eq:linearSGD}. This linearization leads us to provide a much simpler proof for linear convergence analysis of AccSGD by studying the stability of the linear system in Eq. \eqref{eq:linearAccSGD}. We have following Theorem 3 for its convergence condition.

\textbf{Theorem 3.} \emph{Consider the linear system in Eq. \eqref{eq:linearAccSGD}. Assume that $i_t$ is sampled in an IID manner from a uniform distribution, the assumption 1 holds and there exists an equilibrium $x^\ast\in\mathbb{R}^d$ such that $\nabla f(x^\ast)=0$. For a fixed linear convergence rate $0<\rho<\max\left\{1-\frac{0.7^2\xi}{\kappa}, \max_{t}\left\{\frac{\kappa(1-\eta_t \mathbb{L}(x_t))}{\kappa+0.7\xi}\right\}\right\}$, if there exists a $2\times2$ positive definite matrix $P\succ 0$ such that
\begin{align}
\label{eq:condition1AccSGD}
B_t^TPB_t-\rho^2 P\prec0,
\end{align} or the following condition holds
\begin{align}
\label{eq:condition2AccSGD}
0<1-\frac{0.7^2\xi}{\kappa}<\rho \ \ \text{and} \ \ \left(1-\rho\frac{\kappa+0.7\xi}{\kappa}\right)\frac{1}{\mathbb{L}(x_t)}<\eta_t<\frac{1}{\mathbb{L}(x_t)},
\end{align}
then the linear system is exponentially stable, that is, $\small{\left\|\hspace{-0.2cm}
\begin{array}{cc}
m_{t+1}-x^\ast \\
x_{t+1}-x^\ast \\
\end{array}\hspace{-0.2cm}\right\|_2\leq \sqrt{\text{cond}(P)}\rho^t \left\|\hspace{-0.2cm}
\begin{array}{cc}
m_{0}-x^\ast \\
x_{0}-x^\ast \\
\end{array}\hspace{-0.2cm}\right\|_2}$, where $\text{cond}(P)$ is the condition number of $P$.}

\begin{wrapfigure}[15]{r}{2.6in}
\vskip -0.3in
\begin{minipage}{.46\textwidth}
\begin{algorithm}[H]
\caption{PLS-AccSGD.}
\begin{algorithmic}[1]
\State \textbf{input:} $\eta_0$, $\kappa$, $\xi\leq\sqrt{\kappa}$, $\epsilon_1$, $\epsilon_2$.
\State \textbf{initialize:} $x_0$, $u_0=0$, $\alpha=1-\frac{0.7^2\xi}{\kappa}$.
\State \textbf{for} $t=1,\cdots, T-1$ \textbf{do}
\State \hspace{0.4cm} Randomly pick $i_t$ from $\{1,\cdots,n\}$;
\State \hspace{0.4cm} $g_{t}=\nabla f_{i_t}(x_{t})$;
\State \hspace{0.4cm} $L_{i_t}(x_t)=\frac{\|g_{t}-g_{t-1}\|}{\|x_t-x_{t-1}\|+\epsilon_1}$;
\State \hspace{0.4cm} $\eta_t=\eta_0/(L_{i_t}(x_t)+\epsilon_2)$;
\State \hspace{0.4cm} $m_{t+1}=\alpha m_{t}+(1-\alpha)\left(x_{t}-\frac{\kappa\eta_t}{0.7}g_{t}\right)$;
\State \hspace{0.4cm} $x_{t+1}=\frac{0.7}{0.7+(1-\alpha)}\left(x_{t}-\eta_tg_{t}\right)$
\State \hspace{1.5cm} $+\frac{1-\alpha}{0.7+(1-\alpha)}m_{t+1}$;
\State \textbf{end for}
\end{algorithmic}
\label{alg:PLS-AccSGD}
\end{algorithm}
\end{minipage}
\end{wrapfigure}
Theorem 3 shows that the linear convergence conditions in Eqs. \eqref{eq:condition1AccSGD} and \eqref{eq:condition2AccSGD} are simpler than the convergence analysis of AccSGD \cite{Kidambi2018accSGD}. Similar to PLS-AMSGrad, the $2\times2$ LMI condition \eqref{eq:condition1AccSGD} is built by using the control theory and is easily solved by LMI toolbox \cite{Boyd1994lmi}. The condition \eqref{eq:condition2AccSGD} directly opens the learning rate $\eta_t$ is a functional relationship with the local smoothness $\mathbb{L}(x_t)$, $\eta_t=\frac{\eta_0}{\mathbb{L}(x_t)}$, where $\eta_0$ is an initialized learning rate, $1-\rho\frac{\kappa+0.7\xi}{\kappa}<\eta_0<1$. This revePLS $\eta_0$ can be set to a negative value. The reason is that the eigenvalues of the weight matrix in the system \eqref{eq:linearAccSGD} are $1-\frac{0.7^2\xi}{\kappa}$ and $\frac{\kappa(1-\eta_t \mathbb{L}(x_t))}{\kappa+0.7\xi}$. The stability of the system \eqref{eq:linearAccSGD} needs to satisfy the condition \eqref{eq:condition2AccSGD}. Similar to PLS-SGD, $\mathbb{L}(x_t)$ is also sampled uniformly at random from $[1,\cdots ,n]$, that is, $L_{i_t}(x_t)$ is computed by Eq. \eqref{eq:predictlipschitz}. To prevent the over-large learning rate, the learning rate $\eta_t$ is set to $\eta_0/(L_{i_t}(x_t)+\epsilon_2)$ in our PLS-AccSGD, which is outlined in \textbf{Algorithm \ref{alg:PLS-AccSGD}}. Following Remark 2, the $2\times2$ LMI \eqref{eq:condition1AccSGD} is equivalent to the condition \eqref{eq:condition2AccSGD}.
\begin{figure}[t]
\vskip -0.05in
\begin{center}
\centerline{\includegraphics[width=1\columnwidth]{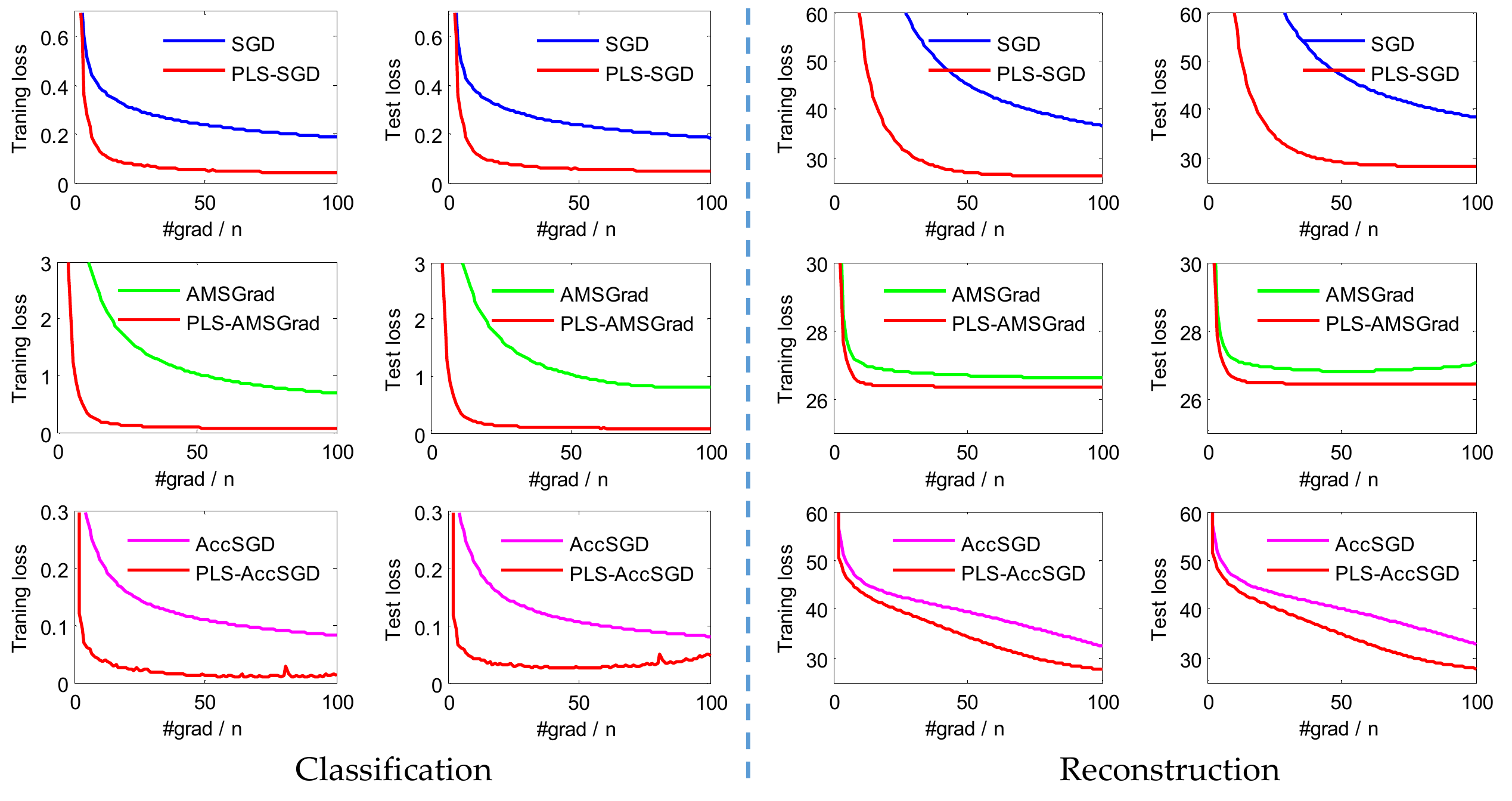}}
\vskip -0.15in
\caption{Performance comparison of SGD, AMSGrad, AccSGD, PLS-SGD, PLS-AMSGrad and PLS-AccSGD on MNIST using neural networks with two fully-connected hidden layers. The left two columns show the training loss and test loss for classification, while right two columns show the training loss and test loss for reconstruction. }
\label{fig:lossresultsonmnist}
\end{center}
\vskip -0.35in
\end{figure}
\hspace{-0.2cm}
\section{Experiments}
In this section, we present empirical results to confirm the effectiveness of the PLS method. For our experiments, we compare PLS-SGD, PLS-AMSGrad and PLS-AccSGD with SGD, AMSGrad and AccSGD by studying the multiclass classification and image reconstruction using neural network with least squares regression (LSR) loss and $\ell_2$-regularization. The weight parameters of the neural network are initialized by using the normalized strategy choosing uniformly from $[-\sqrt{6/(n_{in}+n_{out})},\sqrt{6/(n_{in}+n_{out})}]$, where $n_{in}$ and $n_{out}$ are the numbers of input and output layers of the neural network, respectively. We use mini-batches of size 100 in all experiments.

\textbf{Datasets.} MNIST\footnote{\url{http://yann.lecun.com/exdb/mnist/}} contains 60,000 training samples and 10,000 test samples with 784 dimensional image vector and 10 classes, while CIFA10\footnote{\url{https://www.cs.toronto.edu/~kriz/cifar.html}} includes 50,000 training samples and 10,000 test samples with 1024 dimensional image vector and 10 classes, and 512 dimensional features are extracted by deep residual networks \cite{He2016imresnet} for verifying the effectiveness of our methods.

\textbf{Classification.} We train the neural networks with two fully-connected hidden layers of 500 ReLU units and 10 output linear units to investigate the performance of all algorithms on MNIST and CIFA10 datasets. The $\ell_2$-regularization is $1e\hspace{-0.08cm}-\hspace{-0.08cm}4$ (MNIST) and $1e\hspace{-0.08cm}-\hspace{-0.08cm}2$ (CIFAR10). A grid search is used to determine the learning rate that provides the best performance for SGD, AMSGrad and AccSGD. We set the adaptive learning rate $\eta_t\hspace{-0.08cm}=\hspace{-0.08cm}\eta_0/(L_{i_t}(x_t)+\epsilon_2)$ for PLS-SGD, PLS-AMSGrad and PLS-AccSGD, where $\eta_0$ is chosen as $0.001$ or $0.002$. To enable fair comparison, we set typical parameters $\beta_1\hspace{-0.08cm}=\hspace{-0.08cm}0.9$ and $\beta_2\hspace{-0.08cm}=\hspace{-0.08cm}0.999$ for AMSGrad and PLS-AMSGrad, and set $\kappa\hspace{-0.08cm}=\hspace{-0.08cm}1000$ and $\xi\hspace{-0.08cm}=\hspace{-0.08cm}10$ for AccSGD and PLS-AccSGD \cite{Reddi2018AMSGrad,Kidambi2018accSGD}. For convenience, both parameters $\epsilon_1$ and $\epsilon_2$ are same, $\epsilon=\epsilon_1=\epsilon_2$, and $\epsilon$ chosen as $0.01$ for PLS-SGD and PLS-AMSGrad, and $0.001$ for PLS-AccSGD.

\begin{figure}[t]
\vskip -0.0in
\begin{center}
\centerline{\includegraphics[width=1\columnwidth]{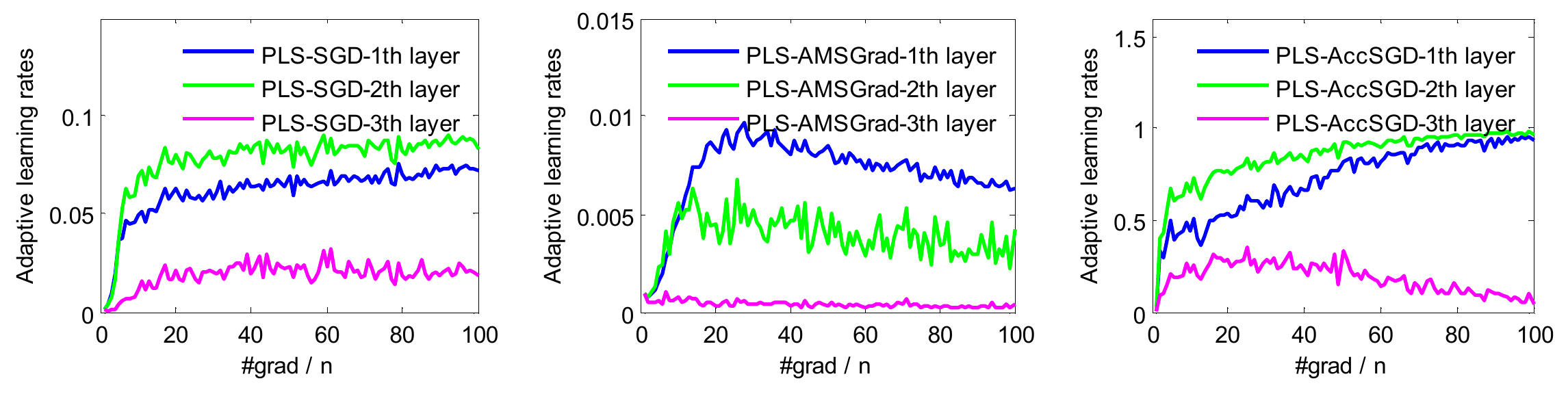}}
\vskip -0.15in
\caption{Adaptive learning rates of three layers of neural networks for classification task on MNIST dataset using PLS-SGD, PLS-AMSGrad and PLS-AccSGD.}
\label{fig:adalearningresultsmnist}
\end{center}
\vskip -0.35in
\end{figure}

We report the training loss and test loss with respect to iterations on MNIST in the left two columns of Figure \ref{fig:lossresultsonmnist}. We can see that PLS-SGD, PLS-AMSGrad and PLS-AccSGD preform much better than SGD, AMSGrad and AccSGD. The important reason is that our PLS method can directly adjust the learning rate from a small initialization value to a suitable value. In practice, a large fixed learning rate results in the explosion of the loss of neural network with ReLU using SGD, AMSGrad and AccSGD since the loss will go to infinity when the learning rate is larger than $0.011$ in our experiments. We observe that the learning rate fast increases in the initial stage and slowly varies in the late stage in Figure \ref{fig:adalearningresultsmnist}. Moreover, there are similar observations on CIFAR10. Due to the limited space, the losses and learning rate are plotted in Figure \ref{fig:resultsoncifar} in the Appendix.

\textbf{Reconstruction.} We train a deep fully connected neural network to reconstruct the images in comparison with all algorithms on MNIST dataset. Its structure is represented as $784\hspace{-0.08cm}-\hspace{-0.08cm}1000\hspace{-0.08cm}-\hspace{-0.08cm}500\hspace{-0.08cm}-\hspace{-0.08cm}200\hspace{-0.08cm}-\hspace{-0.08cm}500
\hspace{-0.08cm}-\hspace{-0.08cm}1000\hspace{-0.08cm}-\hspace{-0.08cm}784$ with the first and last 784 nodes representing the input and output respectively. In this experiment, we provides the best performance for SGD, AMSGrad and AccSGD by searching from a grid learning rates. The initial learning rate $\eta_0$ is set to $5e\hspace{-0.08cm}-\hspace{-0.08cm}7$, $1e\hspace{-0.08cm}-\hspace{-0.08cm}2$ and $1e\hspace{-0.08cm}-\hspace{-0.08cm}7$ for PLS-SGD, PLS-AMSGrad and PLS-AccSGD, respectively. In addition, we also set to $\eta_t=\eta_0/(\sqrt{t}(L_{i_t}(x_t)+\epsilon_2))$ for PLS-AMSGrad, $\beta_1\hspace{-0.08cm}=\hspace{-0.08cm}0.9$, $\beta_2\hspace{-0.08cm}=\hspace{-0.08cm}0.999$, $\kappa\hspace{-0.08cm}=\hspace{-0.08cm}1000$ and $\xi\hspace{-0.08cm}=\hspace{-0.08cm}10$. Moreover, $\epsilon=\epsilon_1=\epsilon_2$, and $\epsilon$ chosen as $0.01$ for PLS-SGD and PLS-AccSGD, and $0.1$ for PLS-AMSGrad.

The training loss and test loss with respect to iterations are reported in the right two columns of Figure \ref{fig:lossresultsonmnist}. We can still see that PLS-SGD, PLS-AMSGrad and PLS-AccSGD have significant better performance than SGD, AMSGrad and AccSGD since our PLS method adaptively adjusts the learning rate to prevent the explosion of the LSR loss with large learning rate. The adaptive learning rate is shown in Figure \ref{fig:lrreconstruct} in the Appendix. We also observe that the learning rate is initialized a small value, fast increases in the earlier stage and slowly varies in the latter stage.


\section{Conclusions}
This paper introduced a predictive local smoothness method for stochastic gradient descent algorithms. This method adjusted automatically learning rate by using the latest gradients to predict the local smoothness. We proposed PLS-SGD, PLS-AMSGrad and PLS-AccSGD algorithms by applying our predictive local smoothness method into the popular SGD, AMSGrad and AccSGD algorithms. We proved that our proposed algorithms enjoyed the linear convergence rate by studying the stability of their transformed linear systems. Moreover, our proof was significantly simpler than the convergence analyses of SGD, AMSGrad and AccSGD. Experimental results verified that the proposed algorithms provided better performance gain than SGD, AMSGrad and AccSGD.

\section{Appendix}

\subsection{Proofs}
\emph{Proof of Theorem 1:} First, we prove that SGD in Eq. \eqref{eq:sgd} is converted into the stochastic linear system in Eq. \eqref{eq:linearSGD}. By putting the Lipschitz linearization in Eq. \eqref{eq:linearizing} into the SGD in Eq. \eqref{eq:sgd}, we have
\begin{align}
\label{eq:sgd1}
x_{t+1}=x_{t}-\eta_t\left(L_{i_t}(x_t)\otimes I_d\right)(x_{t}-x^\ast).
\end{align}
By adding $-x^\ast$ into the both sides of Eq. \eqref{eq:sgd1} and combining like terms, it holds $x_{t+1}-x^\ast=\left(\left(1-\eta_tL_{i_t}(x_t)\right)\otimes I_d\right)(x_{t}-x^\ast)$. Since $i_t$ is sampled in an IID manner from $\{1,\cdots,n\}$, $\mathbb{L}(x_t)=\mathbb{E}\left[L_{i_t}(x_t)\right]=\frac{1}{n}\sum_{i=1}^nL_{i}(x_t)$. By computing the expectation, we thus have Eq. \eqref{eq:linearSGD}.

Second, we construct the Lyapunov function $V(x_t)=(x_t-x^\ast)^T\left(p\otimes I_d\right)(x_t-x^\ast)$, where $p>0$, to prove the stability of the system in Eq. \eqref{eq:linearSGD}. Defining
\begin{align}
\label{eq:deltaVsgd}
\Delta V(x_t)&=V(x_{t+1})-\rho^2V(x_t) \nonumber \\
&=(x_{t+1}-x^\ast)^T\left(p\otimes I_d\right)(x_{t+1}-x^\ast)-\rho^2(x_t-x^\ast)^T\left(p\otimes I_d\right)(x_t-x^\ast) \nonumber \\
&=(x_t-x^\ast)^T\left(\left(1-\eta_t\mathbb{L}(x_t)\right)^2-\rho^2\right)\left(p\otimes I_d\right)(x_t-x^\ast).
\end{align}
Then for any $x_t\neq x^\ast$, $\Delta V(x_t)<0$ if $\left(1-\eta_t\mathbb{L}(x_t)\right)^2-\rho^2<0$, which implies $1-\eta_t\mathbb{L}(x_t)<\rho$. Moreover, $0<1-\eta_t\mathbb{L}(x_t)$. Thus, $0<1-\eta_t\mathbb{L}(x_t)<\rho$, that is, $(1-\rho)\frac{1}{\mathbb{L}(x_t)}\leq \eta_t\leq \frac{1}{\mathbb{L}(x_t)}$. By using the nonnegativity of Eq. \eqref{eq:deltaVsgd}, we have
\begin{align}
(x_{l+1}-x^\ast)^T\left(p\otimes I_d\right)(x_{l+1}-x^\ast)\leq\rho^2(x_l-x^\ast)^T\left(p\otimes I_d\right)(x_l-x^\ast).
\end{align}
Inducting from $l = 1$ to $ t$, we see that for all $t$
\begin{align}
(x_{t}-x^\ast)^T\left(p\otimes I_d\right)(x_{t}-x^\ast)\leq\rho^{2t}(x_0-x^\ast)^T\left(p\otimes I_d\right)(x_0-x^\ast),
\end{align}
which implies $\|x_t-x^\ast\|_2\leq \rho^t \|x_0-x^\ast\|_2$, where $p$ is a positive number. The proof is complete. $\Box$

\emph{Proof of Theorem 2:} First, we prove that AMSGrad is converted into the stochastic linear system in Eq. \eqref{eq:linearAMSGrad}. By putting the Lipschitz linearization Eq. \eqref{eq:linearizing} into Eq. \eqref{eq:AMSGrad}, we have
\begin{subequations}
\begin{align}
\label{eq:AMSGrad5}
m_{t+1}&=\left(\beta_{1t}\otimes I_d\right)m_{t}+(1-\beta_{1t})\left(L_{i_t}(x_t)\otimes I_d\right)(x_{t}-x^\ast),\\
\label{eq:AMSGrad6}
v_{t+1}&=\left(\beta_{2}\otimes I_d\right)v_{t}+(1-\beta_{2})\left(L_{i_t}(x_t)\otimes I_d\right)^2(x_{t}-x^\ast)^2.
\end{align}
\end{subequations}
Because $i_t$ is sampled in an IID manner from $\{1,\cdots,n\}$, $\mathbb{L}(x_t)=\mathbb{E}\left[L_{i_t}(x_t)\right]=\frac{1}{n}\sum_{i=1}^nL_{i}(x_t)$ and $\mathbb{L}^2(x_t)=\mathbb{E}\left[L_{i_t}^2(x_t)\right]=\frac{1}{n}\sum_{i=1}^nL_{i}^2(x_t)$. By computing the expectation, we thus have
\begin{subequations}
\begin{align}
\label{eq:AMSGrad7}
m_{t+1}&=\left(\beta_{1t}\otimes I_d\right)m_{t}+(1-\beta_{1t})\left(\mathbb{L}(x_t)\otimes I_d\right)(x_{t}-x^\ast),\\
\label{eq:AMSGrad8}
v_{t+1}&=\left(\beta_{2}\otimes I_d\right)v_{t}+(1-\beta_{2})\left(\mathbb{L}^2(x_t)\otimes I_d\right)(x_{t}-x^\ast)^2.
\end{align}
\end{subequations}
By adding $-x^\ast$ into the both sides of $x_{t+1}=x_{t}-\eta_t\frac{m_{t+1}}{\sqrt{\widehat{v}_{t+1}}}$ in Eq. \eqref{eq:AMSGrad} and substituting Eqs. \eqref{eq:AMSGrad7} and \eqref{eq:AMSGrad8} into Eq. \eqref{eq:AMSGrad}, it holds
\begin{align}
\label{eq:AMSGrad9}
x_{t+1}-x^\ast&=x_{t}-x^\ast-\eta_t\frac{\beta_{1t}m_{t}+(1-\beta_{1t})\left(\mathbb{L}(x_t)\otimes I_d\right)(x_{t}-x^\ast)}{\sqrt{\overline{v}_{t+1}}}\nonumber \\
&=\left(\frac{\eta_t\beta_{1t}}{\sqrt{\widehat{v}_{t+1}}}\otimes I_d\right)m_{t}+\left(\left(1-\frac{(1-\beta_{1t})\eta_t\mathbb{L}(x_t)}{\sqrt{\widehat{v}_{t+1}}}\right)\otimes I_d\right)(x_{t}-x^\ast),
\end{align}
where $\widehat{v}_{t+1}=\max\{v_{t+1},\widehat{v}_{t}\}$.
So, we have Eq. \eqref{eq:linearAMSGrad} by combining Eq. \eqref{eq:AMSGrad5} with Eq. \eqref{eq:AMSGrad9}.

Second, we construct the Lyapunov function $V(\zeta_t)=\zeta_t^T\left(P\otimes I_d\right) \zeta_t$, where $\zeta_t=\left(\hspace{-0.2cm}\begin{array}{cc}
m_{t} \\
x_t-x^\ast\\
\end{array}\hspace{-0.2cm}
\right)$ and $P\succ 0$ is a $2\times2$ positive matrix, to prove the stability of the system in Eq. \eqref{eq:linearAMSGrad}. Defining
\begin{align}
\label{eq:deltaVAMSGrad}
\Delta V(\zeta_t)&=V(\zeta_{t+1})-\rho^2V(\zeta_t) \nonumber \\
&=\zeta_{t+1}^T\left(A_t\otimes I_d\right)\zeta_{t+1}-\rho^2\zeta_{t}^T\left(A_t\otimes I_d\right)\zeta_{t} \nonumber \\
&=\zeta_{t}^T\left(\left(A_t^TPA_t-\rho^2P\right)\otimes I_d\right)\zeta_{t}.
\end{align}
Then if Eq. \eqref{eq:condition1AMSGrad} is satisfied, then $\Delta V(x_t)<0$ for any $x_t\neq x^\ast$. By using the nonnegativity of Eq. \eqref{eq:deltaVsgd}, we have
\begin{align}
\zeta_{l+1}^T\left(P\otimes I_d\right)\zeta_{l+1}\leq\rho^2\zeta_{l}^T\left(P\otimes I_d\right)\zeta_{l}.
\end{align}
Inducting from $l = 1$ to $ t$, we see that for all $t$
\begin{align}
\zeta_{t}^T\left(P\otimes I_d\right)\zeta_{t}\leq\rho^{2t}\zeta_{0}^T\left(P\otimes I_d\right)\zeta_{0},
\end{align}
which implies $\left\|\hspace{-0.2cm}
\begin{array}{cc}
m_{t+1} \\
x_{t+1}-x^\ast \\
\end{array}\hspace{-0.2cm}\right\|_2\leq \sqrt{\text{cond}(P)} \rho^t \left\|\hspace{-0.2cm}
\begin{array}{cc}
m_{0} \\
x_{0}-x^\ast \\
\end{array}\hspace{-0.2cm}\right\|_2$, where $\text{cond}(P)$ is the condition number of $P$ and $\text{Cond}(A)=\sigma_1(A)/\sigma_p(A)$, where $\sigma_1(A)$ and $\sigma_p(A)$ denote the largest and smallest singular values of the matrix $A$.

Third, we certify the another condition in Eq. \eqref{eq:condition1AMSGrad}. Based on the Lemma 1, $A_t^TPA_t-\rho^2P\prec 0$ is equivalence to $\rho(A_t)<\rho$. The eigenvalues of $A_t$ is calculated by
\begin{align}
\lambda_t I-A_t=\left(\hspace{-0.2cm}
\begin{array}{cc}
\lambda_t-\beta_{1t} & -(1-\beta_{1t})\mathbb{L}(x_t)\\
\frac{\eta_t\beta_{1t}}{\sqrt{\widehat{v}_{t+1}}}& \lambda_t-\left(1-\frac{(1-\beta_{1t})\eta_t\mathbb{L}(x_t)}{\sqrt{\widehat{v}_{t+1}}}\right)\\
\end{array}\hspace{-0.2cm}
\right)=0, \\
(\lambda_t-\beta_{1t})\left(\lambda_t-\left(1-\frac{(1-\beta_{1t})\eta_t\mathbb{L}(x_t)}{\sqrt{\widehat{v}_{t+1}}}\right)\right)+(1-\beta_{1t})\mathbb{L}(x_t)\frac{\eta_t\beta_{1t}}{\sqrt{\widehat{v}_{t+1}}}=0, \\
\lambda_t^2-\left(1+\beta_{1t}-\frac{(1-\beta_{1t})\eta_t\mathbb{L}(x_t)}{\sqrt{\widehat{v}_{t+1}}}\right)\lambda_t+\beta_{1t}=0,\\
\lambda_t=\frac{1+\beta_{1t}-\frac{(1-\beta_{1t})\eta_t\mathbb{L}(x_t)}{\sqrt{\widehat{v}_{t+1}}}\pm \sqrt{\Gamma}}{2},
\end{align}
where $\Gamma=\left(1+\beta_{1t}-\frac{(1-\beta_{1t})\eta_t\mathbb{L}(x_t)}{\sqrt{\widehat{v}_{t+1}}}\right)^2-4\beta_{1t}$.

Similar to the Proof of Proposition 1 \cite{Lessard2016iqc}, if $\Gamma<0$, then the magnitudes of the roots satisfy $|\lambda_t|<\sqrt{\beta_{1t}}<\max_t\{\sqrt{\beta_{1t}}\}$. Then $\Gamma<0$ implies that
\begin{align}
\left(1+\beta_{1t}-\frac{(1-\beta_{1t})\eta_t\mathbb{L}(x_t)}{\sqrt{\widehat{v}_{t+1}}}\right)^2<4\beta_{1t}, \\
-2\sqrt{\beta_{1t}}<1+\beta_{1t}-\frac{(1-\beta_{1t})\eta_t\mathbb{L}(x_t)}{\sqrt{\widehat{v}_{t+1}}}<2\sqrt{\beta_{1t}}, \\
\frac{\left(1-\sqrt{\beta_{1t}}\right)^2\sqrt{\widehat{v}_{t+1}}}{1-\beta_{1t}}\frac{1}{\mathbb{L}(x_t)}<\eta_t<
\frac{\left(1+\sqrt{\beta_{1t}}\right)^2\sqrt{\widehat{v}_{t+1}}}{1-\beta_{1t}}\frac{1}{\mathbb{L}(x_t)}.
\end{align}
The proof is complete. $\Box$

\emph{Proof of Theorem 3:} First, we prove that AccSGD is converted into the stochastic linear system in Eq. \eqref{eq:linearAccSGD}. By putting the Lipschitz linearization in Eq. \eqref{eq:linearizing} into Eq. \eqref{eq:AccSGD}, we have
\begin{subequations}
\begin{align}
\label{eq:AccSGD3}
m_{t+1}&=\left(\alpha \otimes I_d\right) m_{t}+(1-\alpha)\left(x_{t}-\frac{\kappa\eta_t}{0.7}\left(L_{i_t}(x_t)\otimes I_d\right)(x_{t}-x^\ast)\right),  \\
\label{eq:AccSGD4}
x_{t+1}&=\frac{0.7}{0.7+(1-\alpha)}\left(x_{t}-\eta_t\left(L_{i_t}(x_t)\otimes I_d\right)(x_{t}-x^\ast)\right)+\left(\frac{1-\alpha}{0.7+(1-\alpha)}\otimes I_d\right) m_{t+1}.
\end{align}
\end{subequations}
Because $i_t$ is sampled in an IID manner from $\{1,\cdots,n\}$, $\mathbb{L}(x_t)=\mathbb{E}\left[L_{i_t}(x_t)\right]=\frac{1}{n}\sum_{i=1}^nL_{i}(x_t)$. Let $a=\frac{\kappa}{0.7}$ and $b=\frac{1-\alpha}{0.7+(1-\alpha)}$. By computing the expectation, we thus have
\begin{subequations}
\begin{align}
\label{eq:AccSGD5}
m_{t+1}&=\left(\alpha \otimes I_d\right) m_{t}+(1-\alpha)\left(x_{t}-a\eta_t\left(\mathbb{L}(x_t)\otimes I_d\right)(x_{t}-x^\ast)\right),  \\
\label{eq:AccSGD6}
x_{t+1}&=(1-b)\left(x_{t}-\eta_t\left(\mathbb{L}(x_t)\otimes I_d\right)(x_{t}-x^\ast)\right)+\left(b\otimes I_d\right) m_{t+1}.
\end{align}
\end{subequations}
By adding $-x^\ast$ into the both sides of Eq. \eqref{eq:AccSGD5} and Eq. \eqref{eq:AccSGD6}, and substituting Eq. \eqref{eq:AccSGD5} into Eq. \eqref{eq:AccSGD6}, it holds
\begin{subequations}
\begin{align}
\label{eq:AccSGD7}
m_{t+1}-x^\ast&=\left(\alpha \otimes I_d \right)(m_{t}-x^\ast)+\left((1-\alpha)\left(1-a\eta_t\mathbb{L}(x_t)\right)\otimes I_d\right)(x_{t}-x^\ast),  \\
\label{eq:AccSGD8}
x_{t+1}-x^\ast&=\left((1-b)\left(1-\eta_t\mathbb{L}(x_t)\right)\otimes I_d\right)(x_{t}-x^\ast)+\left(b\otimes I_d \right)(m_{t+1}-x^\ast)  \nonumber\\
&=\left(\alpha b\otimes I_d \right)(m_{t}-x^\ast)+\left(\left((1-b)(1-\eta_tL_{i_t}(x_t))+b(1-\alpha)(1-a\eta_t L_{i_t}(x_t))\right)\otimes I_d\right)(x_{t}-x^\ast).
\end{align}
\end{subequations}
Thus, we have the Eq. \eqref{eq:linearAccSGD} by combining Eq. \eqref{eq:AccSGD7} with Eq. \eqref{eq:AccSGD8}.

\begin{figure}[ht]
\vskip -0.0in
\begin{center}
\centerline{\includegraphics[width=1\columnwidth]{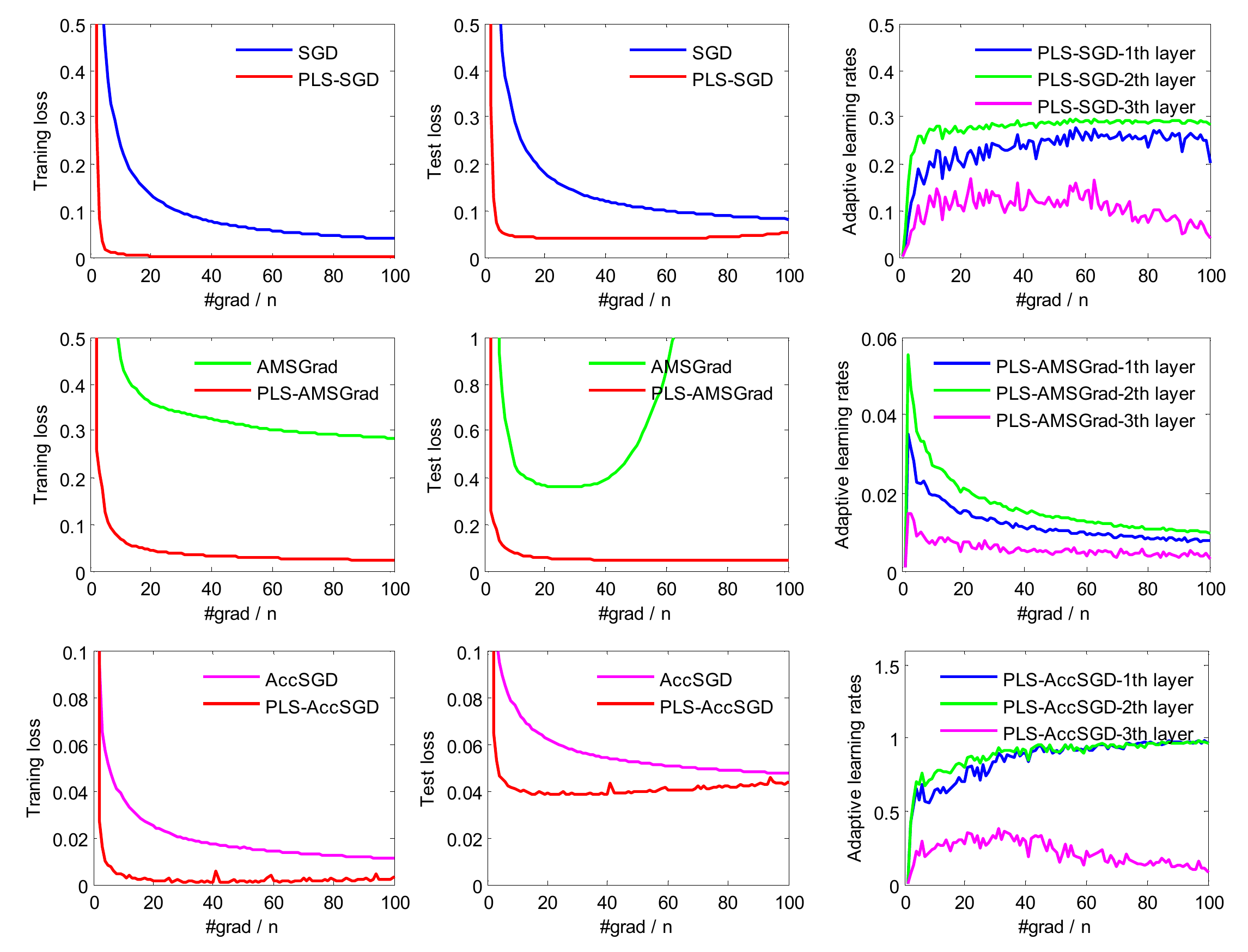}}
\vskip -0.05in
\caption{Performance comparison of SGD, AMSGrad, AccSGD, PLS-SGD, PLS-AMSGrad and PLS-AccSGD on CIFAR10 using neural network with two fully-connected hidden layers. The left, middle, and right columns show the training loss, test loss, and adaptive learning rate, respectively. }
\label{fig:resultsoncifar}
\end{center}
\vskip -0.25in
\end{figure}

\begin{figure}[t]
\vskip -0.0in
\begin{center}
\centerline{\includegraphics[width=1\columnwidth]{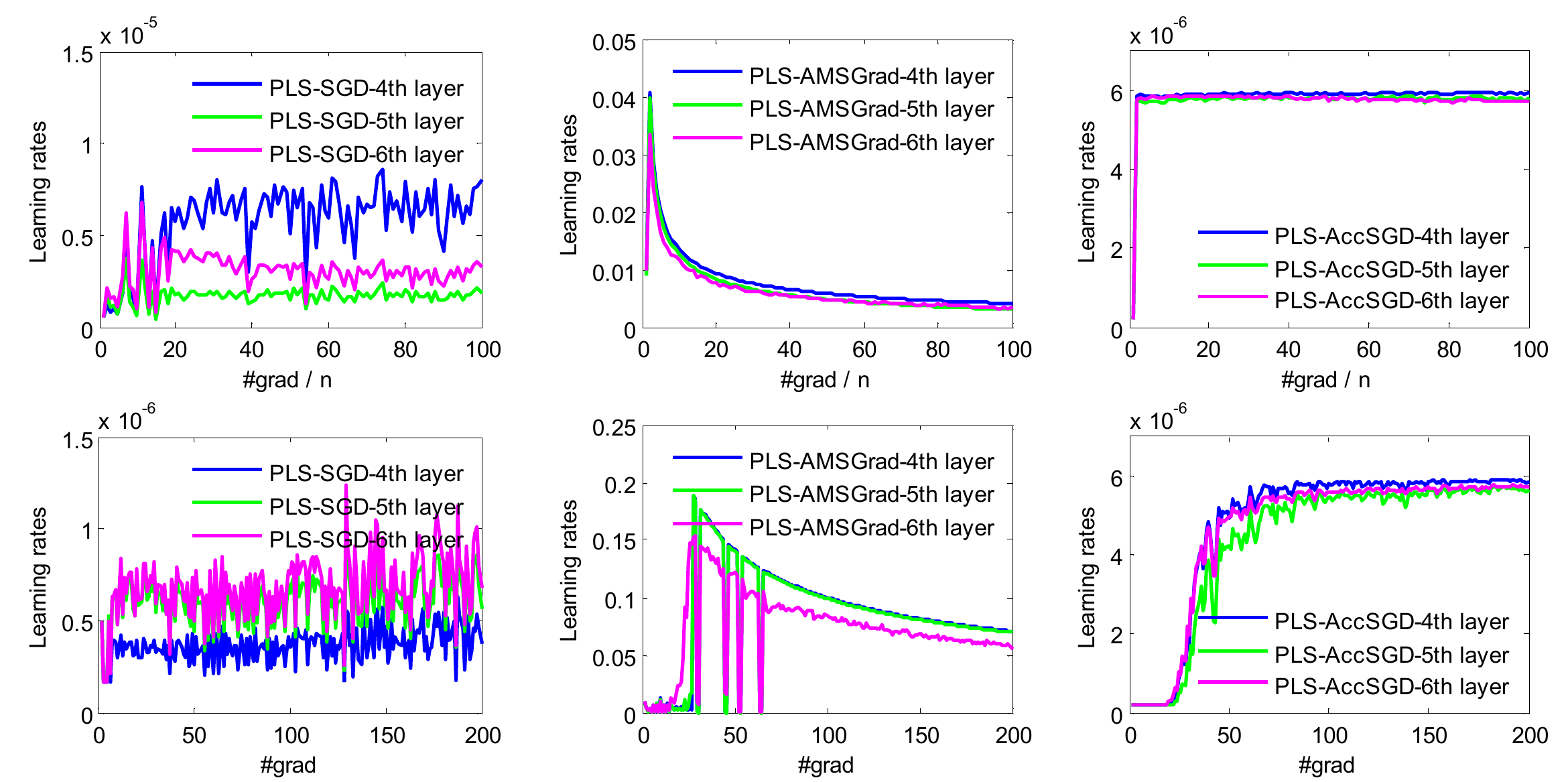}}
\vskip -0.05in
\caption{Adaptive learning rates of different layers of neural network for reconstruction using PLS-SGD, PLS-AMSGrad and PLS-AccSGD on MNIST. The down row shows the adaptive learning rate in the first 200 iterations.}
\label{fig:lrreconstruct}
\end{center}
\vskip -0.25in
\end{figure}

Second, we construct the Lyapunov function $V(\xi_t)=\xi_t^T(P\otimes I_d )\xi_t$, where $\xi_t=\left(\hspace{-0.2cm}\begin{array}{cc}
m_{t}-x^\ast \\
x_t-x^\ast\\
\end{array}\hspace{-0.2cm}
\right)$, $P\succ 0$ is a $2\times2$ positive matrix, to prove the stability of the system in Eq. \eqref{eq:linearAccSGD}. Defining
\begin{align}
\label{eq:deltaVAccSGD}
\Delta V(\xi_t)&=V(\xi_{t+1})-\rho^2V(\xi_t) \nonumber \\
&=\xi_{t+1}^T\left(P\otimes I_d\right)\xi_{t+1}-\rho^2\xi_{t}^T\left(P\otimes I_d\right)\xi_{t} \nonumber \\
&=\xi_{t}^T\left(\left(B_t^TPB_t-\rho^2P\right)\otimes I_d\right)\xi_{t}.
\end{align}
Then if the Eq. \eqref{eq:condition1AccSGD} is satisfied, then $\Delta V(\xi_t)<0$ for any $\xi_t\neq 0$. By using the nonnegativity of Eq. \eqref{eq:deltaVAccSGD}, we have
\begin{align}
\xi_{l+1}^T\left(P\otimes I_d\right)\xi_{l+1}\leq\rho^2\xi_{l}^T\left(P\otimes I_d\right)\xi_{l}.
\end{align}
Inducting from $l = 1$ to $ t$, we see that for all $t$
\begin{align}
\xi_{t}^T\left(P\otimes I_d\right)\xi_{t}\leq\rho^{2t}\xi_{0}^T\left(P\otimes I_d\right)\xi_{0},
\end{align}
which implies $\small{\left\|\hspace{-0.2cm}
\begin{array}{cc}
m_{t+1}-x^\ast \\
x_{t+1}-x^\ast \\
\end{array}\hspace{-0.2cm}\right\|_2\leq \sqrt{\text{cond}(P)}\rho^t \left\|\hspace{-0.2cm}
\begin{array}{cc}
m_{0}-x^\ast \\
x_{0}-x^\ast \\
\end{array}\hspace{-0.2cm}\right\|_2}$, where $\text{cond}(P)$ is the condition number of $P$.

Third, we certify the another condition in Eq. \eqref{eq:condition2AccSGD}. Based on the Lemma 1, $B_t^TPB_t-\rho^2P\prec 0$ is equivalence to $\rho(B_t)<\rho$.
The eigenvalues of $B_t$ is calculated as follows. By adding a product of $-b$ and the first row of $B_t$ into the second row of $B_t$, $B_t$ is rewritten as $\widehat{B}_t$:
\begin{align}
\widehat{B}_t=\left(\hspace{-0.2cm}
\begin{array}{cc}
\alpha & (1-\alpha)(1-a\eta_t \mathbb{L}(x_t))\\
0 & (1-b)(1-\eta_t\mathbb{L}(x_t))\\
\end{array}\hspace{-0.2cm}
\right), \\
\lambda_t I-\widehat{B}_t=\left(\hspace{-0.2cm}
\begin{array}{cc}
\lambda_t-\alpha & -(1-\alpha)(1-a\eta_t\mathbb{L}(x_t))\\
0& \lambda_t-(1-b)(1-\eta_t\mathbb{L}(x_t))\\
\end{array}\hspace{-0.2cm}
\right)=0.
\end{align}
The two eigenvalues of $B_t$ is
\begin{align}
\lambda_{1t}=\alpha, \ \ \ \ \lambda_{2t}=(1-b)(1-\eta_t\mathbb{L}(x_t)).
\end{align}
Since $\lambda_{1t}>0$ and $\lambda_{2t}>0$, we have
\begin{align}
\label{eq:eigenvaluesB}
0<\lambda_{1t}=\alpha<\rho, \ \ \ \ 0<\lambda_{2t}=(1-b)(1-\eta_t\mathbb{L}(x_t))<\rho.
\end{align}
By substituting $\alpha=1-\frac{0.7^2\xi}{\kappa}$ and $b=\frac{1-\alpha}{0.7+(1-\alpha)}$ into Eq. \eqref{eq:eigenvaluesB}, it holds the condition in Eq. \eqref{eq:condition2AccSGD}. The proof is complete. $\Box$

\subsection{Figures}

In the classification experiment, we select the learning rate from $\{0.011, 0.009 0.008, 0.007, 0.006, 0.05, 0.004\}$ for providing the best performance of the SGD, AMSGrad and AccSGD algorithms. In the reconstruction experiment, the learning rate is chosen from $\{6,5,4,3,\}e\hspace{-0.08cm}-\hspace{-0.08cm}7$ for SGD and AccSGD and $\{10,7,5,3\}e\hspace{-0.08cm}-\hspace{-0.08cm}2$ for AMSGrad due to the explosion of the LSR loss with large learning rate, and select the learning rate from these sets for providing the best performance of the algorithms. To prevent the over-fitting, the learning rate $\eta_t$ is set to $\eta_0/\sqrt{t}$ for AMSGrad (except MNIST).

Figure \ref{fig:resultsoncifar} reports the adaptive learning rate, the training loss and test loss with respect to iterations on CIFAR10 for classification, while Figure \ref{fig:lrreconstruct} shows the adaptive learning rate with respect to iterations on MNIST for reconstruction.
\medskip

\bibliographystyle{plainnat}
\bibliography{example_nips} 

\end{document}